\documentclass[letterpaper, 10 pt, conference]{ieeeconf}  % Comment this line out if you need a4paper

\IEEEoverridecommandlockouts% This command is only needed if 
\usepackage{subfig}
\usepackage{graphicx}
\usepackage{comment}
\usepackage{booktabs}
\usepackage{diagbox}
\usepackage{multirow}
\usepackage{colortbl}
\usepackage{makecell}
\usepackage{float}
\usepackage{placeins}
\usepackage{stfloats}
\usepackage{amsmath} % assumes amsmath package installed
\usepackage[bb=px]{mathalfa} 
\usepackage[colorinlistoftodos]{todonotes}

\renewcommand{\arraystretch}{1.2}

\title{\LARGE \bf
Learning to Double Guess: An Active Perception Approach for Estimating the Center of Mass of Arbitrary Objects
}

\author{Shengmiao Jin, Yuchen Mo, Wenzhen Yuan$^{1}$% <-this % stops a space
\thanks{$^{1}$ University of Illinois Urbana-Champaign}
\thanks{\{\tt\small jin45, yuchenm7, yuanwz\}@illinois.edu}%
}

\begin{document}

\maketitle
\thispagestyle{empty}
\pagestyle{empty}

%%%%%%%%%%%%%%%%%%%%%%%%%%%%%%%%%%%%%%%%%%%%%%%%%%%%%%%%%%%%%%%%%%%%%%%%%%%%%%%%
\begin{abstract}

Manipulating arbitrary objects in unstructured environments is a significant challenge in robotics, primarily due to difficulties in determining an object's center of mass. This paper introduces U-GRAPH: Uncertainty-Guided Rotational Active Perception with Haptics, a novel framework to enhance the center of mass estimation using active perception. Traditional methods often rely on single interaction and are limited by the inherent inaccuracies of Force-Torque (F/T) sensors. Our approach circumvents these limitations by integrating a Bayesian Neural Network (BNN) to quantify uncertainty and guide the robotic system through multiple, information-rich interactions via grid search and a neural network that scores each action. We demonstrate the remarkable generalizability and transferability of our method with training on a small dataset with limited variation yet still perform well on unseen complex real-world objects.

\end{abstract}

%%%%%%%%%%%%%%%%%%%%%%%%%%%%%%%%%%%%%%%%%%%%%%%%%%%%%%%%%%%%%%%%%%%%%%%%%%%%%%%%

\section{Introduction}

%\wenzhen{Opening is not very relevant and not well connected with the content}
%\wenzhen{Somewhere in the introduction you need to briefly review the state of the current research on the topic}
 %\wenzhen{Lots of things in this paragraph is not very relevant}
With the growing interest in robotics manipulation in the wild, researchers have been investigating ways for robots to interact with different objects. A key factor in achieving a secure grasp is the proximity of the grasp point to the object's center of mass (CoM). Yet, not much research has been trying to address a generalizable approach to estimate the CoM of arbitrary objects. Much recent research on grasping has been focused solely on grasping onto the object \cite{kosmose24} \cite{yuchen23}, often overlooking how the choice of grasping points and the physical properties of objects affect successful manipulation. It has been noted in works like \cite{Poseit} and \cite{rotate_stability} that improper grasping points or poses on objects with certain properties can lead to failures, typically due to rotational and incidental slips. Knowing the CoM of the object, on the other hand, can help decide a more stable and robust grasping strategy, and ensure safety during manipulation.

\begin{figure}[htbp]
\vspace{-5mm}
\begin{center}
\includegraphics[scale=0.31]{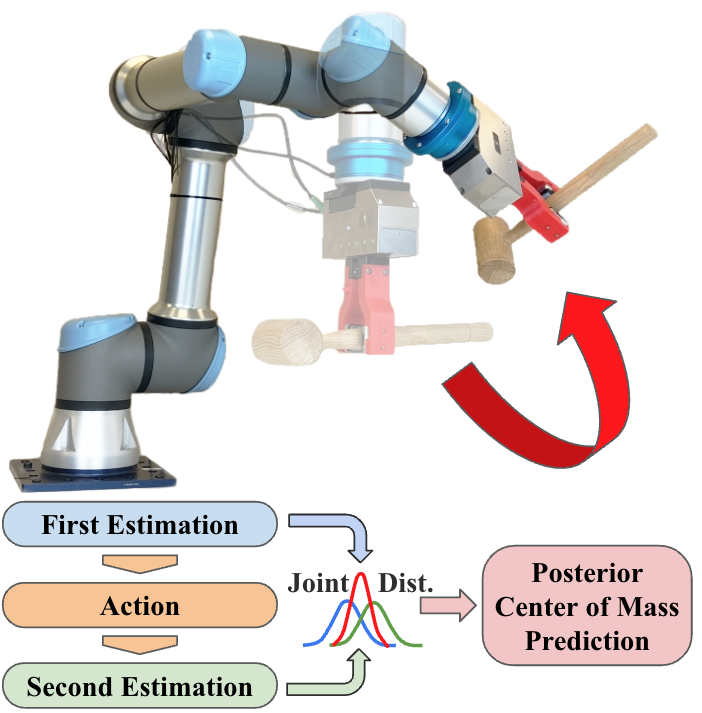}
\end{center}
\caption{We design an active perception algorithm to estimate the center of mass of arbitrary objects. Our algorithm uses the first estimation from the F/T reading to infer a new rotational orientation that improves the estimation, then executes the action and estimates again with a second F/T reading. }
\label{teaser}
\vspace{-5mm}
\end{figure}

This work provides a framework for perceiving the CoM of an arbitrary object. object. The most intuitive solution is using the Force-Torque (F/T) sensor's reading and solving CoM analytically. However, the physical limitations of F/T sensors provide inaccurate readings caused by the measurement noise and in-hand slips, therefore analytical solutions often fail. Additionally, analytical solutions fails to provide uncertainty measurements, making it impossible to estimate an informed action to improve measurements. Instead, we turn to a data-driven solution to learn the correlation between the F/T Reading and the CoM locations with neural networks. Moreover, a vertical grasping pose prevents obtaining the location of CoM on the z-axis. Hence, we cannot predict precisely enough with a single F/T reading. Our strategy incorporates active perception techniques to enhance accuracy, which identifies the orientation that maximizes information gained from the initial grasp. Our methodology involves actively guiding the robot to adjust its orientation which will provide a more accurate prediction. 

Therefore, we propose \textbf{U-GRAPH}: Uncertainty-Guided Rotational Active Perception with Haptics to address the difficulty found in the CoM Estimation problems as illustrated in Fig. \ref{teaser}. We construct a Bayesian Neural Network (BNN) \cite{1992bnn} for uncertainty quantification, obtaining a standard measurement deviation aligned with physical intuition. First, the robot picks up the object with a fixed orientation, providing a prior estimation of the CoM. With the estimation's mean and standard deviation, we employ ActiveNet to infer the orientation that can provide the greatest information gain. From the subsequent orientation, the BNN can provide a new estimation of CoM from the latest F/T reading. 

Our system is trained on 204 grasp trials each with 100 rotations with 20 weight distributions. Despite being trained only on two customized simple data collection objects, we demonstrate its ability to generalize to any arbitrary rigid object in real life. It shows an average of 1.47-centimeter error with a 7.6\% error on unseen complex real-world objects in a zero-shot transfer manner. Our system shows generalizability to all kinds of objects with different contact geometry, surface friction, and overall shapes. %With the result from CoM estimation, the robot can interact with any object more safely, preventing failure caused by slips. Our method shows the potential to apply to any non-visual properties that are needed for object manipulation. %Our results with CoM estimation indicate a step towards more intelligent robotic manipulation with complex objects.% In summary, the key contribution of this work is the active perception algorithm that can perform robust and accurate center of mass estimation on arbitrary objects with a small and limited dataset.
%\wenzhen{A few sentence to talk about the future: like how the methodology can be developed or used in other tasks? How the result of CoM measurement can help robots?}

\section{Related works}
%\wenzhen{You need to have a subsection to talk about CoM estimation}
\subsection{Physical Property Estimation}
%\wenzhen{I don't think this topic is very relevant. It's not wrong to keep it, but if you need to save space you can remove this}
A critical premise for the successful manipulation of different objects is an understanding of their physical properties. For example, liquid properties were well-studied by previous research \cite{joe-liquid} \cite{xiaofeng-solid} \cite{matl}. Moreover, many papers also showed that with perception, the precision and accuracy of manipulation can be increased \cite{stir-to-pour} \cite{joe-kitchen} \cite{pancake24}. 
Rigid body properties are also important for manipulation. Zeng et al. used residual physics to help decide a better-tossing policy for a wide range of objects \cite{zeng2019tossingbot}. Wang et al. demonstrated their algorithm can learn implicit properties and improves their swing policy with tactile explorations \cite{swingbot}. Murooka et al. embedded physical reasoning into manipulation skills \cite{murooka_physics}. Most previous works only consider physical property as a vague or distilled representation, while we focus on estimating the explicit measurement of the center of mass.  
\subsection{Center of Mass Estimation}
A more related set of works is directly aimed at estimating the center of mass. Hyland et al. utilized iterative pushing to find a 2-dimensional CoM \cite{com23}. McGovern et al. pointed out that with reinforcement learning of stacking random shape objects in a simulator, they can estimate the CoM \cite{com19}. McGovern and Xiao also proposed a Reinforcement Learning pipeline in the real world to estimate the CoM of utensils with torque-sensing \cite{com22}. However, to our knowledge, none of the methods are generalizable to find the 3-dimensional CoM for arbitrary objects in real life, which is the problem we are trying to solve. 

\subsection{Active Perception}
%\wenzhen{The best way to write the related work is to emphasize the connection of those works with your work. Active perception is very wide. In your work, you want to highlight the algorithms for active perception, and explain how your method is new, so in the review you should talk more about their contribution in the algorithm}
Humans naturally possess the ability to explore an object actively with touch \cite{activetouch}. Inspired by this, many works have studied active perception in robotics with haptic or tactile sensing. 
Xu et al. employed active tactile perception to classify objects and showed improvement in both accuracy and efficiency \cite{xu23tandem3d}. Uttayopas et al. utilized active haptic sensing to classify objects with different properties \cite{Utta23haptic}. Kuzliak et al. designed a framework to interactively learn the physical properties of an object with informed action selection \cite{kruzliak2024interactivelearningphysicalobject}. The works mentioned above only concern a discrete action space, but in our problem setting, we have a continuous action space. Yuan et al. used active perception to decide the next best grasping location to help increase accuracy for clothes material classification \cite{clothes}. Ketchum used active exploration on a scene to understand its property and shows that with active planning the exploration can performed better \cite{biotacHaptic24}. In our work, on the other hand, instead of classification, we are interested in regression tasks with continuous action space and aim to solve the CoM estimation with only two actions.

\section{Method}
%\wenzhen{In general I think this section is unclear. You need to explain specifically your pipeline of estimating CoM. The definition of the poses to grasp objects is also part of the method}

%\wenzhen{start with a high-level overview of your method -- assuming people forgot most of things in introduction}
Targeting a generalized and robust CoM estimation framework, we propose U-GRAPH: Uncertainty-Guided Rotational Active Perception with Haptics. This system incorporates a BNN that processes 6-dimensional force-torque readings and 2-dimensional orientation data to yield a 3-dimensional CoM estimation. U-GRAPH also features ActiveNet, which utilizes the output from the BNN to determine the next best action. Assuming that the robot has already grasped the object, we perform two measurements at different orientations to accurately estimate its CoM. The BNN supplies both prior predictions and quantifies uncertainty through the standard deviation. The ActiveNet takes in prior estimation distribution and uses grid search to calculate a score for each action to determine the best one. Specifically, the action space is the 2-dimensional orientation of the grasping pose. We define the action executed as changing the pose. In the subsequent subsections, we discuss an intuitive physics model, introduce the individual modules of this framework, and present the implementation of online inference.

%To obtain a robust and generalizable estimation of the center of mass, we propose U-GRAPH: Uncertainty-Guided Rotational Active Perception with Haptic.
%U-GRAPH features a Bayesian Neural Network with mean estimation and uncertainty quantification to provide both the prior and secondary estimation of the CoM of the grasping object from the Force Torque reading. This network is trained with a 6-dimensional force-torque reading input, and 2-dimensional orientation input, with the supervised label as the 3-dimensional center of mass. We also believe that this can be generalized to other physical property problems with various input and output modalities. Our system also includes a second part that focuses on determining the next best action. We refer to this part as the ActiveNet, which trains from the result of the first networks, with data of the same label but went through different actions. In our setup, the action is the 2-dimensional rotation orientation of the grasping pose. The second part can also be substituted for any continuous action space. A visualization of the system can be found in Fig. \ref{active}, and we will explain the two parts in detail in the following subsections.
\subsection{Intuitive Model of Arbitrary Object's Center Of Mass}
\label{model}
\begin{figure}[htbp]
\begin{center}
\includegraphics[scale=0.4]{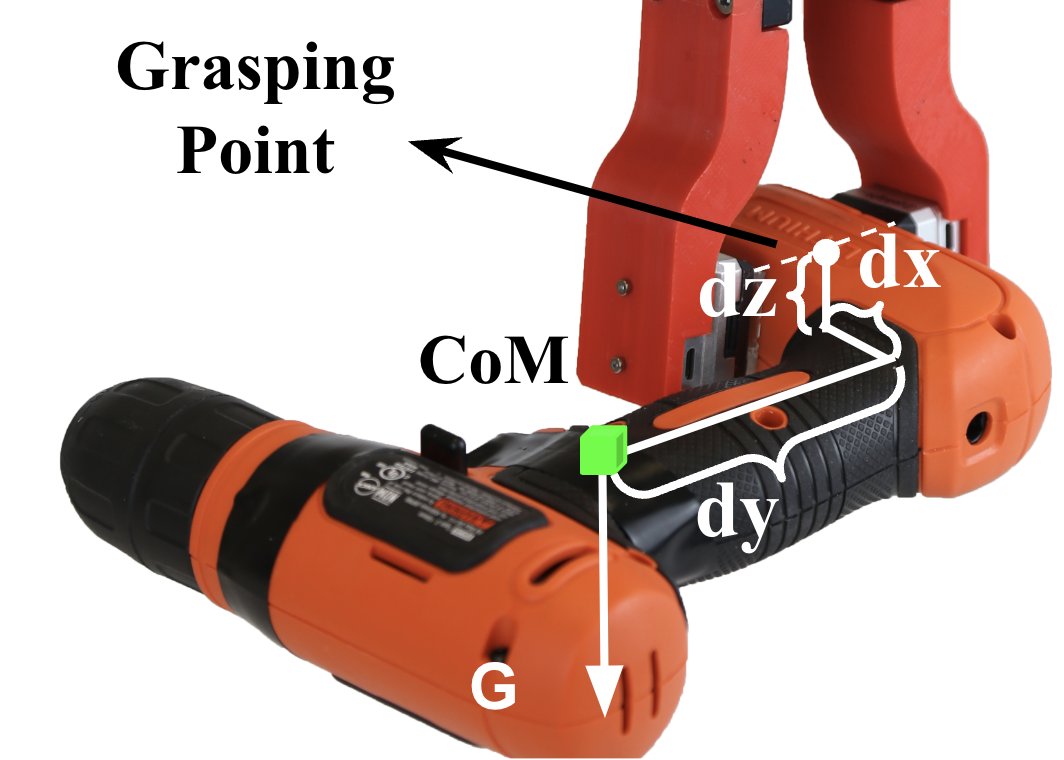}
\end{center}
\caption{Illustration of the simplified model of CoM on a real-world object. In our setup, we try to estimate the $\mathrm{d}x$, $\mathrm{d}y$, and $\mathrm{d}z$ which are the displacement of CoM away from the grasping point.}
\label{model_figure}
\vspace{-3mm}
\end{figure}

%\wenzhen{Shoudl this be in the method section?}
After grasping the object, we define its CoM by some displacement $dx$, $dy$, and $dz$ away from the grasping point. These axes are defined in the world coordinates, as illustrated in Fig. \ref{model}.
Ideally, we could directly employ an analytical solution using the 6-dimensional F/T reading from an F/T sensor to determine the CoM. However, real-world complications, such as measurement noise and potential in-hand slipping of the object, complicate this process. To counteract these issues, a second measurement is necessary. %This follow-up allows for recalibration and adjustment based on any discrepancies noted from the initial data, providing a more accurate and reliable estimation of the CoM. 
Our method aims to reduce the effect of real-world challenges towards a more robust and accurate prediction.

\begin{figure*}[htbp]
\vspace{2mm}
\begin{center}
\includegraphics[scale=0.26]{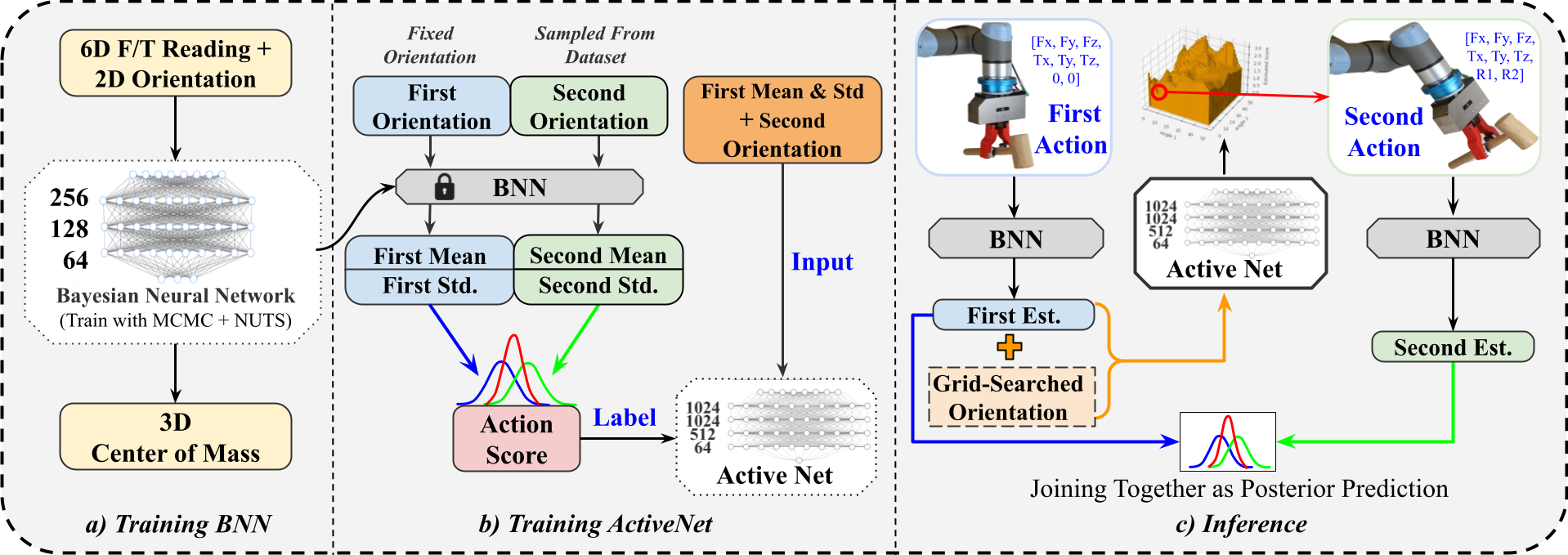}
\end{center}
\caption{a) Flowchart for training Bayesian Neural Network. We train BNN with Markov Chain Monte Carlo and No U-Turn Sampler iteratively. b) Flowchart for training an active perception module. We calculate the score from two orientations as the supervised label of the ActiveNet. We use the first prediction's mean and standard deviation along with the second angle as the input to the network. c) Flowchart for inferencing with U-GRAPH. The robot first grasps with a fixed orientation, then passes the F/T reading with (0, 0) as orientations into the BNN. We use ActiveNet and grid search to find the second action. We pass the second F/T reading with the orientation through BNN to get a secondary prediction and join that with the first prediction to form the posterior prediction.}
%\wenzhen{This caption is too long}}
\label{active}
\vspace{-3mm}
\end{figure*} 

\subsection{Bayesian Neural Network for Uncertainty Qualification}
%\wenzhen{It's unclear how this section is relevant to your method. At least somewhere you need to explain you are using this network to estimate CoM}
%\wenzhen{Start your subsection with directly what you are doing here, or what problem you are trying to solve. Don't start with the background or motivation. Check with other method sections too}

%\wenzhen{It's not clear what do you want to show in the paragraph. It's always nice to start with a sentence for explaining what you want to show, and then goes to details. }
%A common approach to data-driven solutions is to use a Neural Network such as a Multi-Layer Perceptron (MLP) to regress from a large dataset. However, in our scenario, the challenge extends beyond mere predictions; we also aim to quantify the uncertainty associated with these predictions. For this purpose, we utilize a Bayesian Neural Network (BNN), which maintains a similar structure to a traditional MLP but operates under non-deterministic principles. The variability in these outputs allows us to determine the uncertainty of predictions, quantifying it as the standard deviation of the estimated center of mass.

The purpose of using BNN is to get a standard deviation for its output value. Instead of training to specify the exact weight of each network node, in the BNN framework, 
we want to learn a posterior distribution $p(w|D)$ given the input dataset $D$. Each node in our BNN will have a distribution instead of a deterministic value. Given this distribution, we can obtain the estimated distribution of unseen data     $P(\hat{\mathbf{y}}|\hat{\mathbf{x}})$ by getting the expectation of the predictive distribution: $P(\hat{\mathbf{y}}|\hat{\mathbf{x}}) = \mathbb{E}_{P(\mathbf{w}|\mathcal{D})}[P(\hat{\mathbf{y}}|\hat{\mathbf{x}},\mathbf{w})]$, $\mathbf{w}$ denotes the posterior distributions of the nodes in the BNN, $\hat{\mathbf{x}}$ denotes the input testing data and $\hat{\mathbf{y}}$ denotes the output prediction.

However, to evaluate this expectation value, we will need an infinite ensemble of networks as mentioned in \cite{weightuncertainty}. To practically approximate this, Monte Carlo sampling methods, particularly Markov Chain Monte Carlo (MCMC), are employed to reduce training and inference costs. MCMC provides unbiased samples from the posterior, facilitating effective posterior inference and backpropagation.  
Further improving this approach, the Hamiltonian Monte Carlo (HMC) and No U-Turn Sampler (NUTS) are incorporated to avoid the inefficient random walk behavior and dynamically determine the optimal number of steps in the HMC. This automatically adjusts the BNN parameters after each sample to enhance convergence and accuracy \cite{hoffman2011nouturnsampleradaptivelysetting} \cite{Brooks_2011}. 

To implement the BNN and MCMC with NUTS, we used Pyro \cite{pyro} to construct the network, train on our dataset, and evaluate its predictive function. This method gives us reliable uncertainty of the regression prediction of our MLP for active perception. The illustration of the BNN training process can be found in Fig. \ref{active} a).

\subsection{ActiveNet: Action Selection Network}
As mentioned before, our actions have 2 degrees of freedom, the last two joints on the robot are free to move, while all other joints are fixed during perception.
We always keep the orientation [0,0] as the first orientation. This is the orientation where the gripper points straight down, as shown in Fig. \ref{active} c).
To find the best second orientation that improves the prediction result, we consequently design ActiveNet and use grid search to find such orientation. The most intuitive way to generate a new action is to directly estimate from the prediction of the BNN and train the network to predict the best subsequent orientation. In our case, there are usually multiple orientations that the robot can take to minimize the error of CoM prediction. A simple regression model explicitly predicts a single ``best" action, but it can overlook other ``good" actions, especially if these are localized away from the highest peak. 

We therefore try to perform a grid search through the action space and estimate a score to determine how good each action is. For simplicity, we define this score as the error of estimation obtained by the BNN after we perform a specific rotation that results in the orientation $a$. 
%During data collection, we did not collect data explicitly according to the grid we used for data collection, and our grid size (2500) is much larger than our dataset size (100), proving that our network can do robust interpolation of data points for score estimation. 
As a result, the input of our ActiveNet as illustrated in Fig. \ref{active} b) has three parts, the estimation from BNN, the standard deviation from BNN, and a new proposed action that be scored on. 
The output of the ActiveNet is a score of this new proposed action. %For simplicity, we defined this as the error of the mean of the joint prior and posterior distribution after the second orientation. Both estimations are obtained from the pre-trained BNN.
%During training, the proposed action and ground truth action score is from a randomly selected action, and its error when passed through a pre-trained BNN. 

\subsection{Inference}
Our inference pipeline is illustrated in Fig. \ref{active} c). We first use the fixed orientation to generate a prior estimation of the CoM location. Then ActiveNet performs a grid search over the entire action space and calculates the score for each action with prior estimation as input. It uses the action with the minimum action score to proceed. %\wenzhen{The following part is not about action selection. You should use a new sub-section, or merge it with the overview, or find some other ways to accommodate it.}
After we obtain the new F/T reading from the second orientation, we then predict the CoM again with the same BNN. Finally, we treat each orientation as an independently observed measurement of CoM. Since our network can provide a quantified uncertainty, we assume the two measurements are Gaussian. Therefore we can obtain the posterior estimation with: $$
%with the following formula:$$
\mu_{final} =\frac{ (\frac{\mu_1}{\sigma_1^2} + \frac{\mu_2}{\sigma_2^2})}{(\frac{1}{\sigma_1^2} + \frac{1}{\sigma_2^2})}$$
%\yuchen{$\sigma_{final}$?}

\section{Experiment Setup}

In this section, we discuss the hardware setup of the CoM estimation problem. We also explain how to set up the hardware, collect training data, and implement models.

\subsection{Hardware Setup}
\label{hardware}
\begin{figure}[htbp]
\vspace{2mm}
\begin{center}
\includegraphics[scale=0.36]{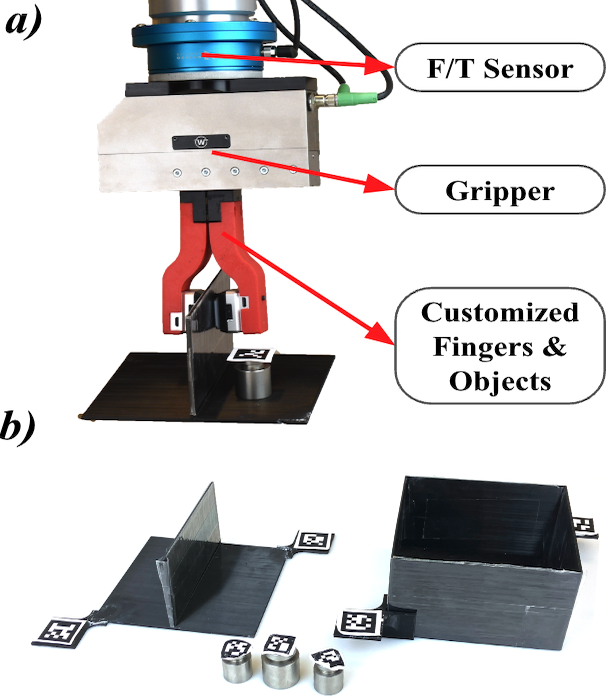}
\end{center}
\caption{a) Example of a data collection robot grasping with the location of F/T Sensor and gripper. b) Printed data collection objects in the real world, and standard lab weights for training data collection. AprilTags are placed on all of the objects. We refer to the object on the left as Plate and the object on the right as Box. }
\label{setup}
\vspace{-5mm}
\end{figure}

The hardware system features a 6-DoF UR5e robot arm. Attached to the robot's wrist is a 6-axis NRS-6050-D80 F/T sensor from Nordbo Robotics with a sampling rate of 1000 Hz. The arm is also equipped with a WSG-50 2-fingered gripper from Weiss Robotics with customized 3D-printed PLA fingers. The system is shown in Fig. \ref{setup} a).

For data collection, we designed and 3D-printed two objects with dimensions of 15cm $\times$ 15cm $\times$ 8cm, each including two holders sized 4cm $\times$ 4cm for placing AprilTags \cite{olson2011tags}. The plate object weighs 127.36 grams and allows grasping onto the center. The box object weighs 185.36 grams and is designated to be grasped on the side. We utilize standard laboratory weights for the experiments, specifically two 100-gram weights and one 200-gram weight. Fig. \ref{setup} b) shows the printed version of these objects, as well as weights that are randomly placed on them. 

\subsection{Data Collection}
%\wenzhen{I think you should highlight the experiment design part-- the creation of those standard objects and your design thoughts are very interesting. You can mention this in the introduction. Also in this section you should emphasize it. E.g. separate the data collection part; add it in the subsection title or the opening sentence}

As mentioned in Sec. \ref{hardware}, we only collect data from the two customized objects for CoM estimation. Our model is based on the premise that the CoM of any grasped object can be fundamentally described by the offsets $\mathrm{d}x$, $\mathrm{d}y$, $\mathrm{d}z$, and the gravitational force $G$ acting on the object. During each trial, we randomly select 0 to 2 weights to be fixed onto one of the objects. On the software side, our data collection algorithm first uses the overhead camera to detect AprilTag and calculate the CoM while figuring out the graspable zone on the object. It randomly generates a valid grasping point from the graspable zone and calculates the $dx$, $dy$, and $dz$ from the CoM. Finally, the robot moves to the location and grasps the object to start a trial of data collection. This data collection algorithm saves the trouble of intensive human labor and allows us to do a larger scale of data collection. 
%\yuchen{``This'' refers to?}

After securing a grasp, the robot rotates the object to 100 different orientations (excluding the default [0,0]), recording the F/T readings at each position. We use the difference between the F/T reading with the object gripped and the F/T reading with nothing between fingers. Then, we loop through the entire 100 different orientations to calculate the action score for each of them. In total, we spent about 150 hours on data collection to generate a dataset from 204 different grasps, comprising 18,893 F/T readings. 

%For the active perception part, we assume the first action is fixed. We use the rotational orientation of $[0,0]$ as the fixed first orientation, which is the orientation that the gripper points straight down, as shown in Fig. \ref{active}. ActiveNet utilizes the variety of rotational data collected from identical grasps but differing orientations to train for better predictive performance. During the evaluation and testing phases, we employ a grid search across a 50 by 50 grid to comprehensively explore a rotational space spanning $2\pi$ by $1\pi$ radians. This method allows us to systematically assess the potential reorientation angles, ensuring optimal grasp and manipulation strategies are developed based on the CoM predictions from the BNN.
\subsection{Model Implementation}
The BNN has a backbone with a hidden size of [256, 128, 64]. To speed up the training process, we first train with PyTorch for a deterministic MLP with RAdam \cite{liu2019radam} optimizer and a learning rate of 0.001 for 500 epochs. Then we use the pre-trained weights as mean and standard deviation of 0.5 as the initialization for our BNN. We train the BNN using Pyro with 1000 samples and 200 warmup steps. 

The ActiveNet is a 5-layer MLP with a hidden size of [1024, 1024, 512, 64]. We train the ActiveNet with RAdam optimizer with a learning rate of 0.0001 for 500 epochs.
\section{Results and Discussion}

In this section, we present the result obtained from the CoM estimation with the U-GRAPH pipeline. We evaluated the performance of our model on the customized data collection setup with unseen weight distributions. In addition, we experiment on real-world objects with known CoM to validate the effectiveness of our proposed framework. 

\subsection{Baseline Methods}%\wenzhen{Will it be better to put this in IV?}
%\vspace{-1mm}
In contrast to our proposed method, we implement the following 3 different baseline methods:

\textbf{Analytical Solution}: The analytical solution assumes a perfect-world scenario with no noise in the F/T measurement and no in-hand slip. We can easily obtain the CoM of any object using the following formula: $\Vec{r}_{CoM} = \frac{\Vec{\tau} \times \Vec{F}}{|\Vec{F}|^2}$,
where $\Vec{F}$ denotes the force reading and $\Vec{\tau}$ denotes the torque readings. Since we only use the $[0,0]$ orientation, no torque should be caused by offsetting the $Z$-axis. However, real-world measurement noise introduces randomness into this calculation. Additionally, the F/T sensor produces inefficiently accurate torque measurement when the sensor is not placed vertically. Therefore, in this paper, we use analytical solutions only in the default pose as the baseline method.

% \yuchen{Explain why this gives random value on z-axis: in raw data corresponding torque should be 0 but in practice some small noise}

\textbf{One Grasp}: The One Grasp method only uses the first part of our proposed pipeline which is the model that takes in one F/T measurement and tries to infer the CoM. Unlike the analytical solution, this method also uses a neural network for estimation. The MLP used is the same one we used for active perception inference. %This method serves as an ablation study of using a second action

The previous two baselines have a fundamental flaw that based on one grasp it is impossible to evaluate the offset on the Z-axis.
% \yuchen{So is the z value from these methods simply random?}
% \samuel{sort of}

\textbf{Random Rotate}: The random rotation method uses two measurements similar to our proposed method. Instead of an informed action, this method uses a random action selected from the continuous action space to perform the measurement again. After getting the new reading, we will use the same BNN and joint distribution methods as our proposed method to estimate the CoM. %This method serves as an ablation study to use an informed action.

% \yuchen{Table 1 axis should be in world frame}
% \yuchen{Shouldn't OOD stand for Out-of-Distribution?}

\subsection{CoM Estimation on Customized Training Objects}

\begin{figure}[hbtp]
\vspace{2mm}
\begin{center}
\includegraphics[scale=0.23]{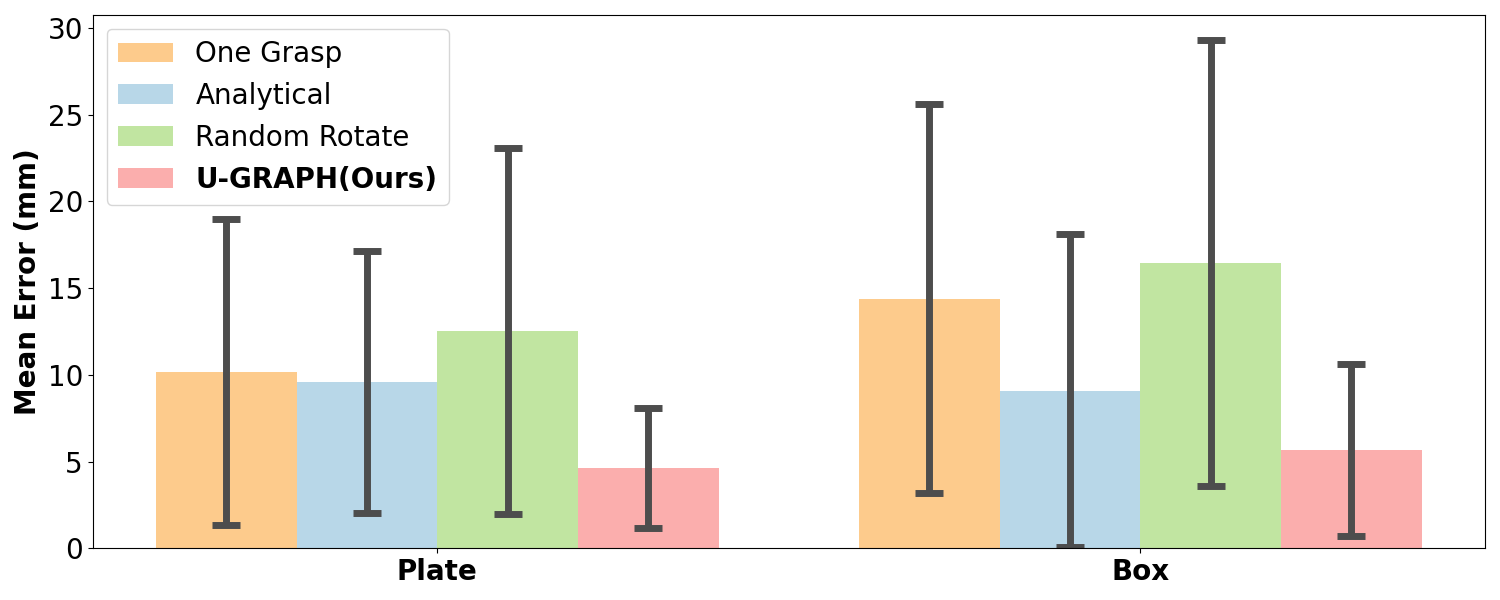}
\end{center}
\caption{Mean error and mean standard deviation (shown with the error bar) of the estimated center of mass for customized objects obtained from different methods.}
\label{result1}
\vspace{-5mm}
\end{figure}

Our first experiment used the same plate and box setup but varied weight configurations. We tested five different weight configurations across both objects: no weight, a single 100-gram weight, two 100-gram weights placed together, two 100-gram weights placed separately, and two separate weights that weigh 300 grams in total. For each configuration, we performed five randomly selected grasps.  The data for this experiment are captured using the same overhead camera and AprilTags setup as training. The results of these experiments are detailed in Figure \ref{result1}, where we present a comprehensive analysis including the mean error and the mean standard deviation for each method applied.

\subsection{CoM Estimation on Unseen Real-World Objects}
We also performed experiments on a set of 12 real-world objects that are commonly seen in daily life. We predefined the grasping point and found the ground truth CoM by balancing the object on each axis with a gripper.
% \yuchen{Sounds not trivial to me}
The objects have weights ranging from 43.4 grams to 613.2 grams with maximum dimensions from 56 mm to 285 mm. We try to create variations on the $X$, $Y$, and $Z$ axes of the measurement to assess the robustness of the methods. We present the result on the error of each axis for each object in Tab. \ref{table2} along with the dimension and weight of each object. We will give a more comprehensive discussion and analysis in Sec. \ref{discuss}.
\renewcommand{\arraystretch}{1.1}
\begin{table*}[btp]
\centering
\footnotesize
\vspace{2mm}
\begin{tabular}{m{14mm}|m{15mm}*{6}{|c@{\hspace{2mm}}c@{\hspace{2mm}}c@{\hspace{2mm}}}} % Adjusted spacing after Z columns

\multirow{3}{=}{\raisebox{5mm}{\centering Objects}}
& \multicolumn{1}{c|}{\raisebox{4mm}{\centering Image}} & \multicolumn{3}{c|}{\centering \includegraphics[width=11mm]{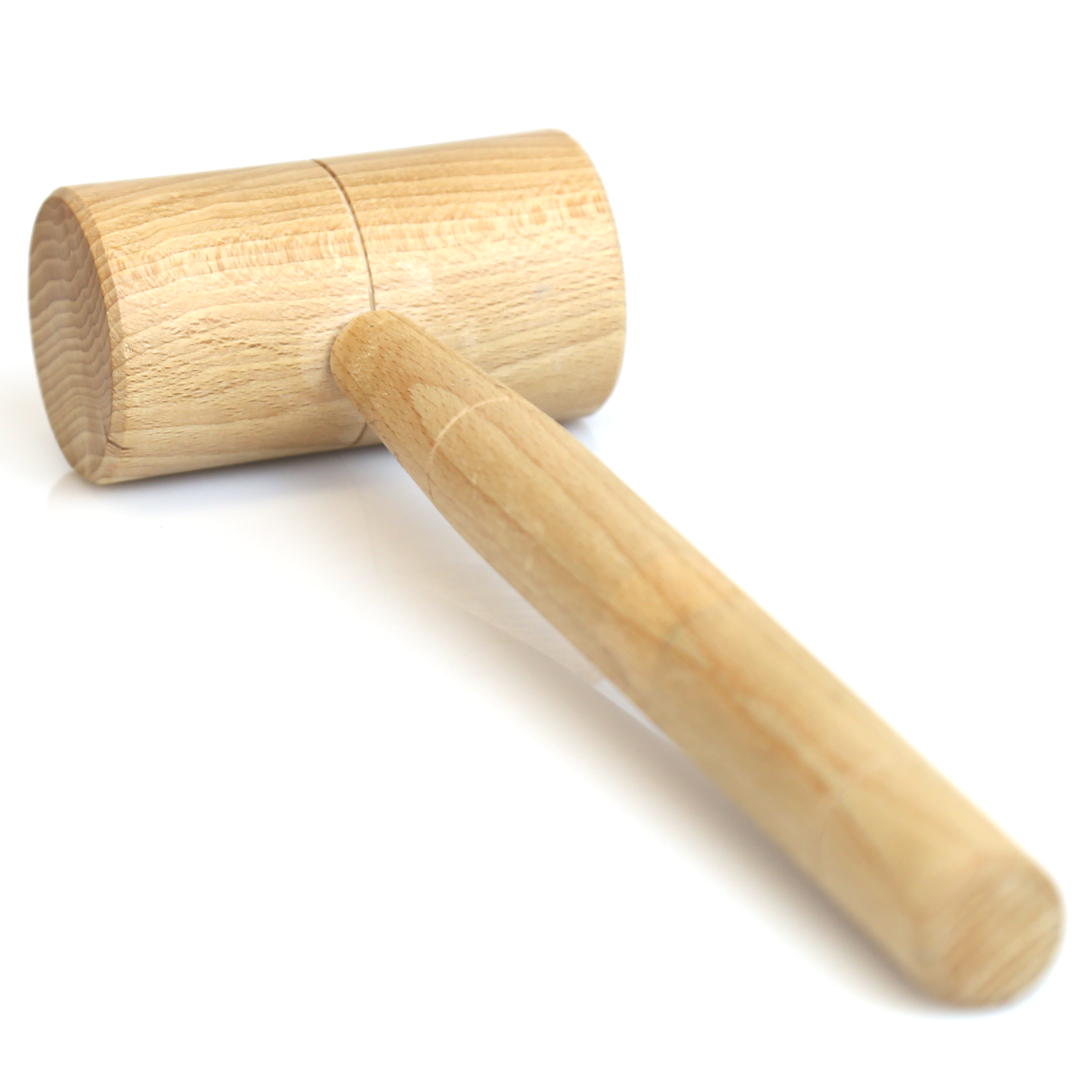}}& \multicolumn{3}{c|}{\centering \includegraphics[width=11mm]{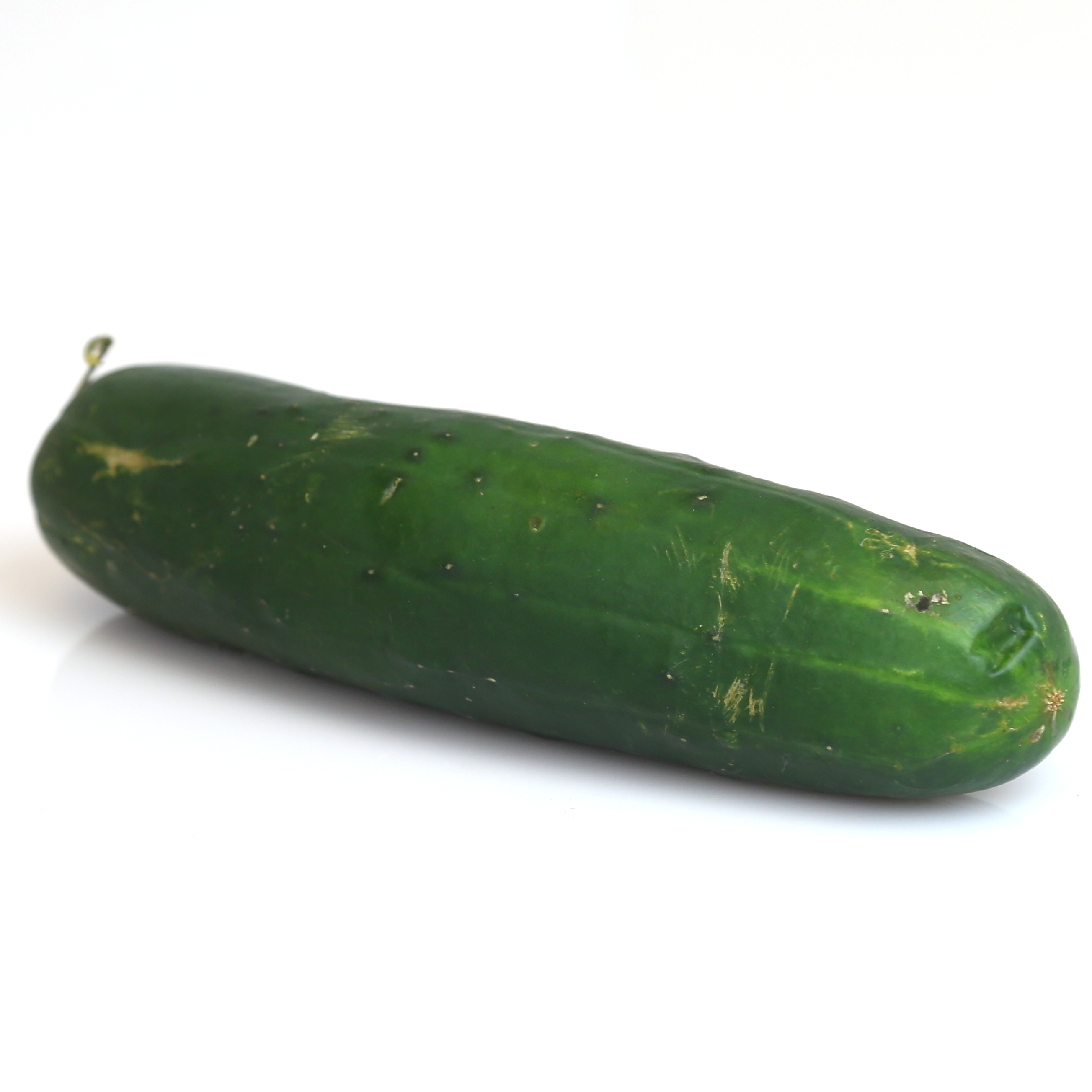}} & \multicolumn{3}{c|}{\centering \includegraphics[width=11mm]{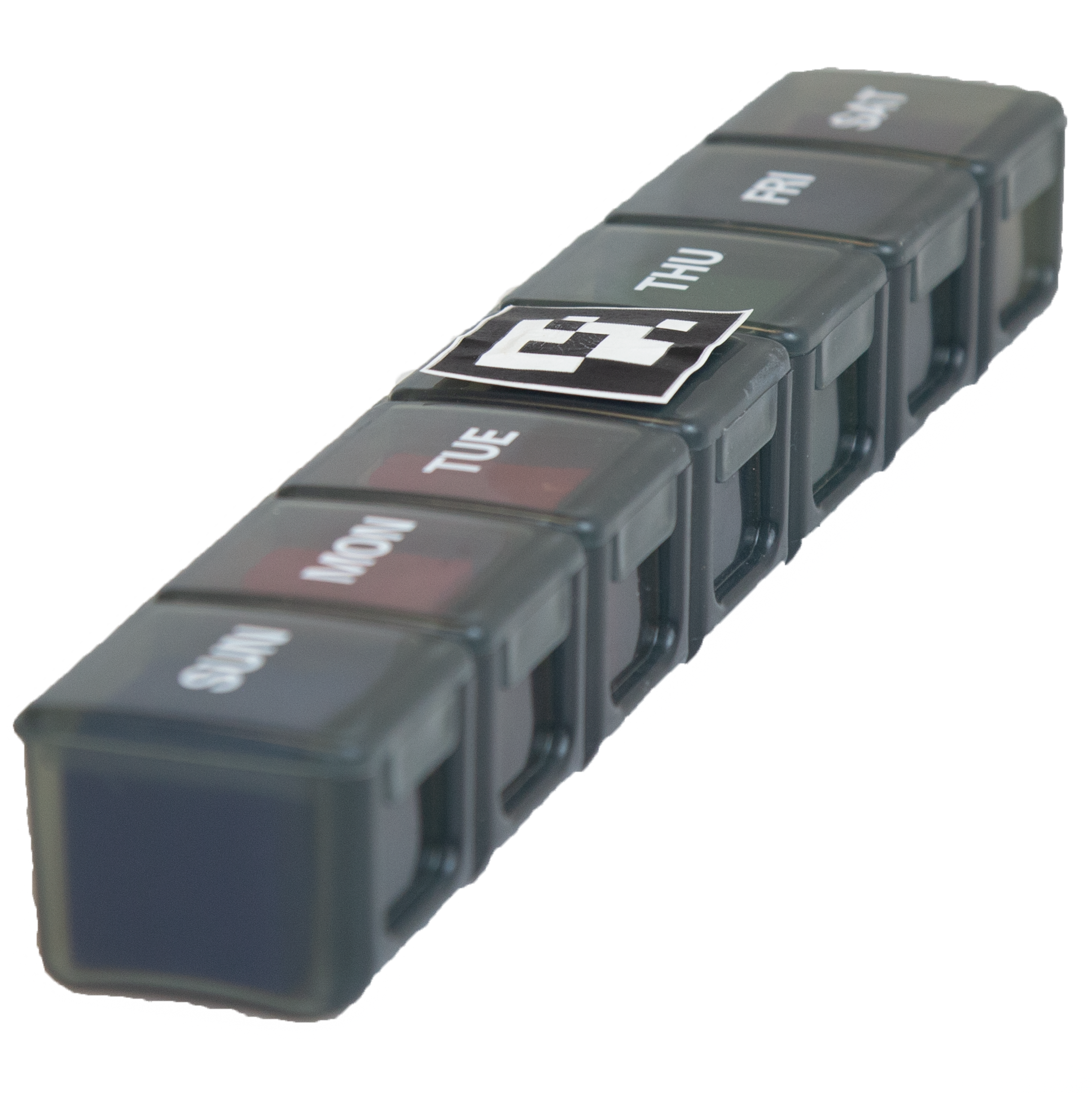}}& \multicolumn{3}{c|}{\centering \includegraphics[width=11mm]{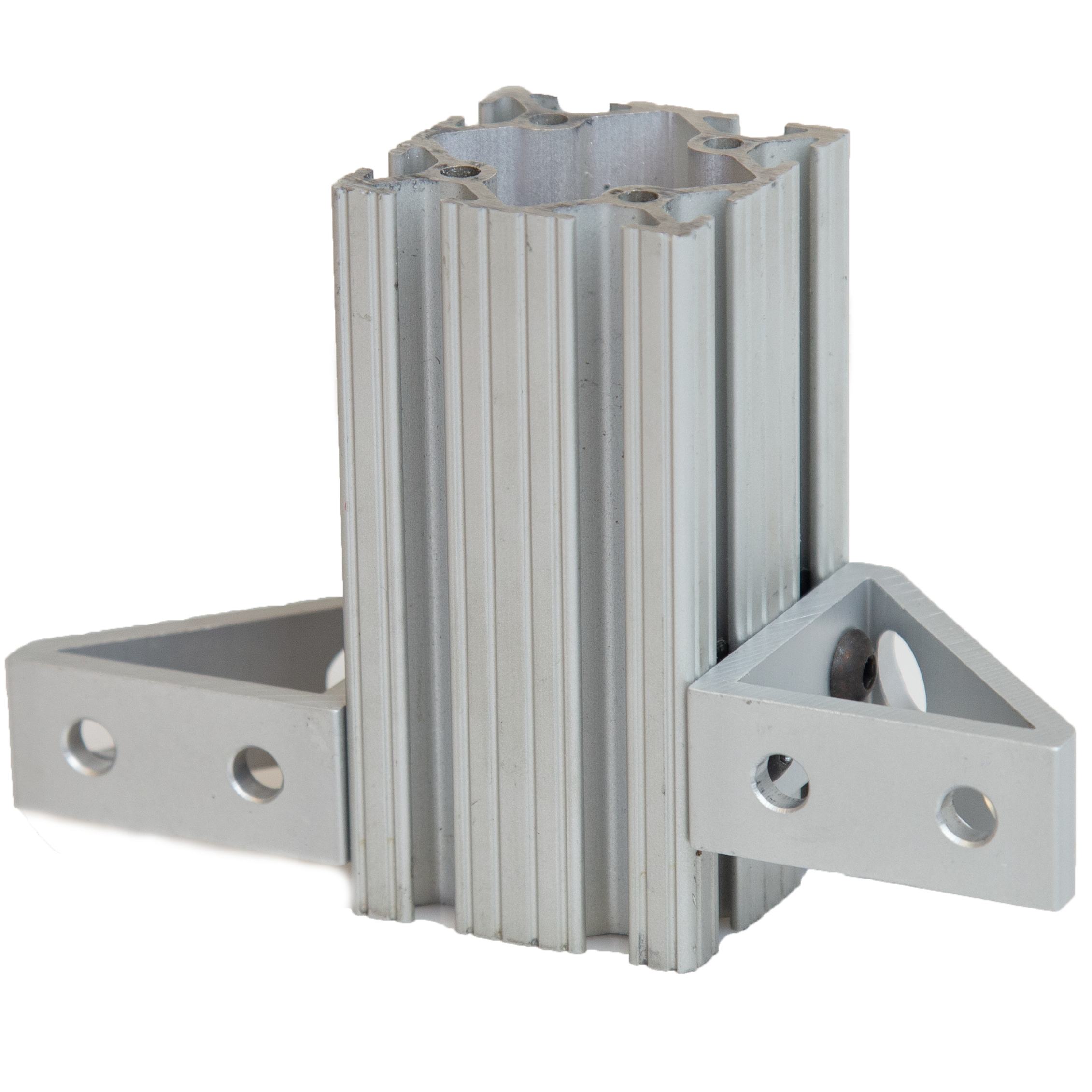}}& \multicolumn{3}{c|}{\centering \includegraphics[width=11mm]{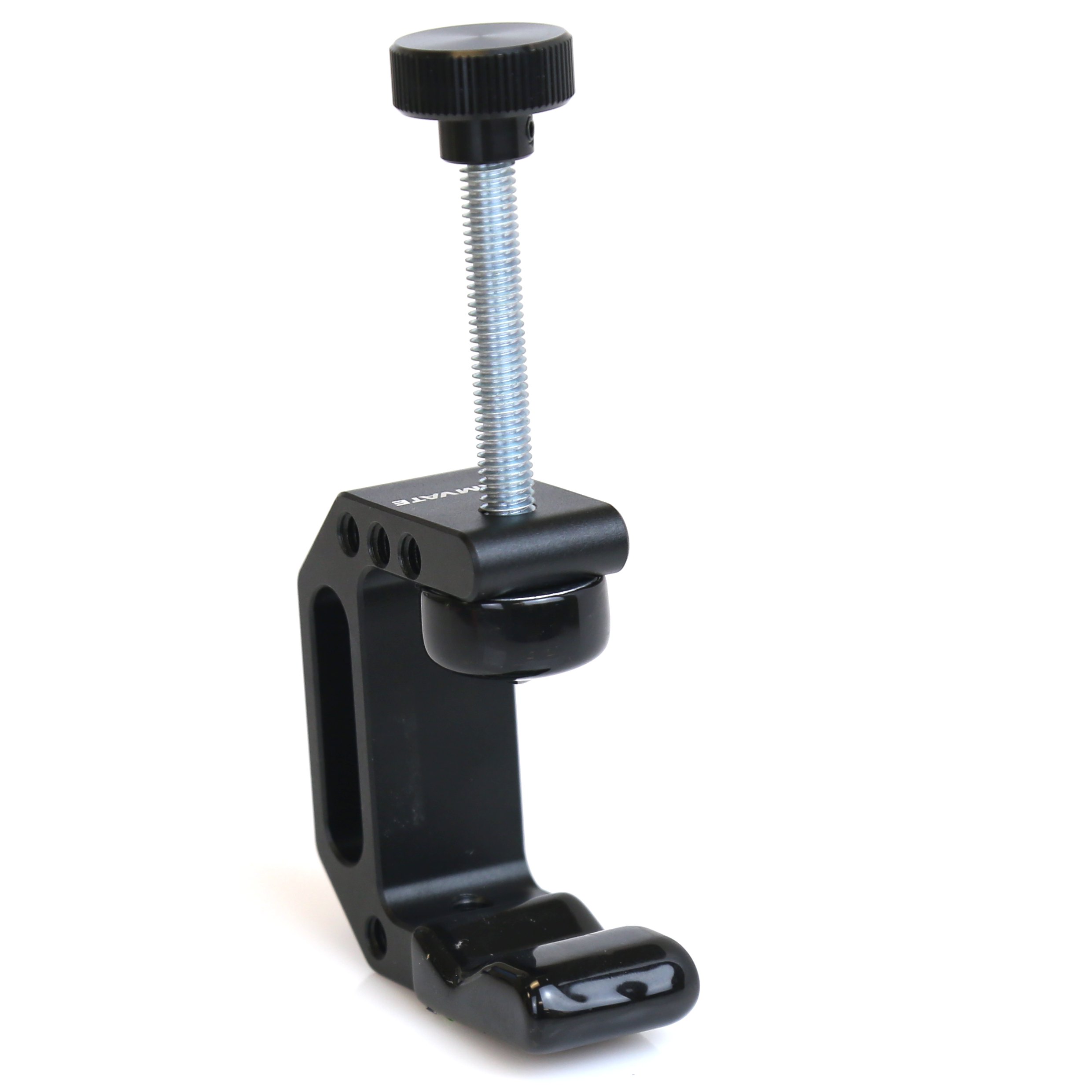}}& \multicolumn{3}{c}{\centering \includegraphics[width=11mm]{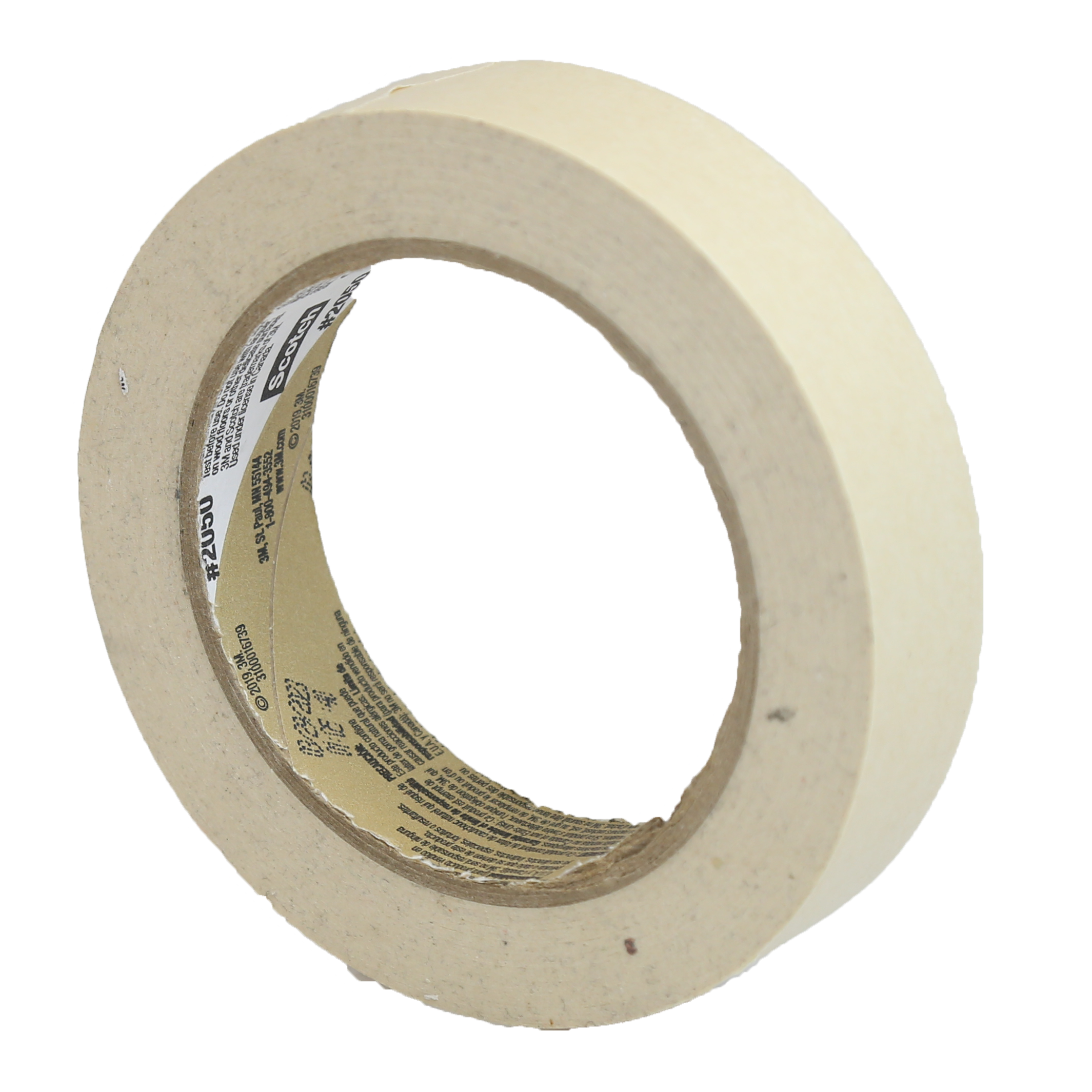}} \\

&\centering Dimensions (mm) & \multicolumn{3}{c|}{285 $\times$ 50 $\times$ 50}& \multicolumn{3}{c|}{185 $\times$ 45 $\times$ 40}& \multicolumn{3}{c|}{244 $\times$ 42 $\times$ 39}& \multicolumn{3}{c|}{51 $\times$ 152 $\times$ 99}& \multicolumn{3}{c|}{140 $\times$ 65 $\times$ 25}& \multicolumn{3}{c}{105 $\times$ 105 $\times$ 25}\\
& \centering Weight (g) & \multicolumn{3}{c|}{260.7} & \multicolumn{3}{c|}{325.1}& \multicolumn{3}{c|}{136.9}& \multicolumn{3}{c|}{307.7}& \multicolumn{3}{c|}{144.8}& \multicolumn{3}{c}{122.7}\\
 \hline
 \multirow{5}{10mm}{\newline \newline \newline \newline Prediction Error (mm)} &
\centering Axis & X & Y & Z & X & Y & Z & X & Y & Z & X & Y & Z & X & Y & Z  & X & Y & Z \\ 
\cline{2-20}
& \centering One Grasp Only & 11.4 & 14.2 & 14.0 & 17.8 & 16.1 & 6.2 & 12.0 & 5.9 & 8.4 &16.8&16.9&19.8& 8.1 &16.7&11.5&2.6&9.1&18.4  \\
& \centering Analytical Solution&6.9&5.5&8.4&6.3&\textbf{4.7}&17.9&14.0&3.1&9.3&6.1&3.8&10.7&9.9&5.5&19.4&2.3&\textbf{3.2}&21.2 \\
& \centering Random Rotate&10.1&22.2&13.7&13.6&18.8&\textbf{3.3}&12.2&4.9&13.4&\textbf{5.2}&14.8&17.1&13.3&11.2&9.2&2.4&9.4&18.5\\
 & \centering \textbf{U-GRAPH (Ours)}&\textbf{2.7}&\textbf{4.0}&\textbf{5.6}&\textbf{5.0}&5.6&4.3&\textbf{3.9}&\textbf{2.2}&\textbf{6.3}&\textbf{5.2}&\textbf{3.0}&\textbf{9.3}&\textbf{3.3}&\textbf{3.0}&\textbf{5.0}&\textbf{2.2}&6.8&\textbf{7.9}\\
\multicolumn{1}{c}{\newline }\\
\multirow{3}{=}{\raisebox{5mm}{Objects}}
& \multicolumn{1}{c|}{\raisebox{5mm}{Image}} & \multicolumn{3}{c|}{\centering \includegraphics[width=12mm]{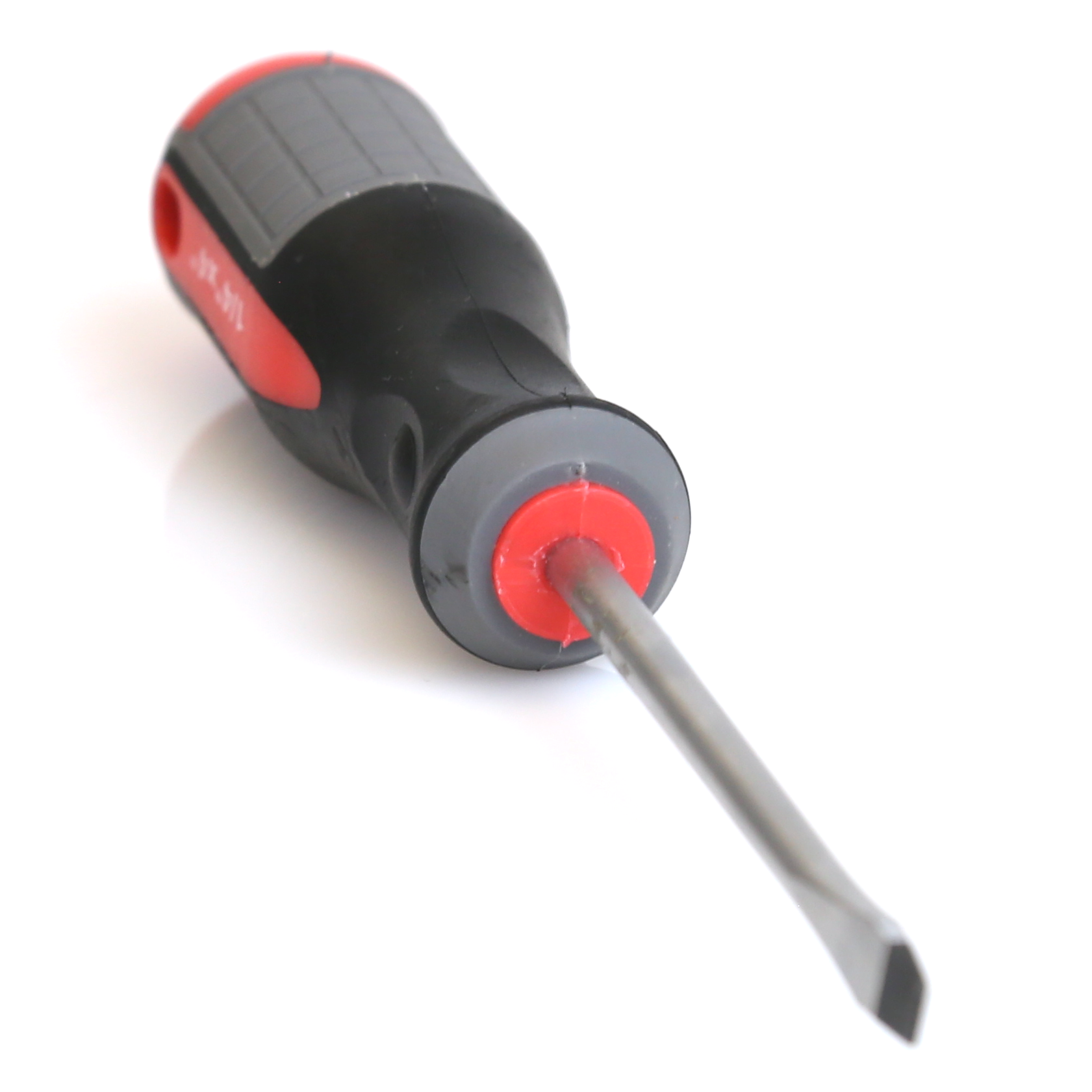}}& \multicolumn{3}{c|}{\centering \includegraphics[width=12mm]{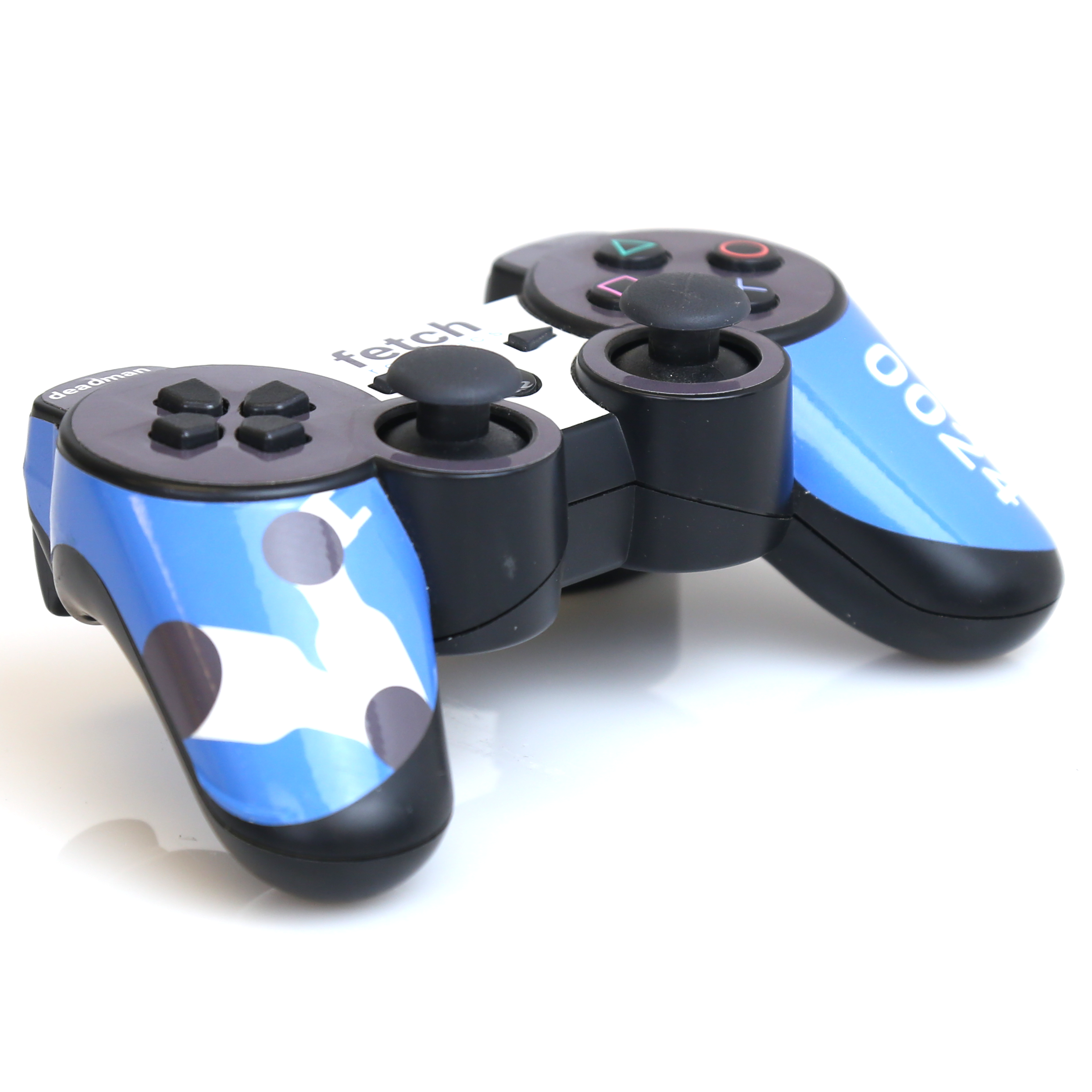}} & \multicolumn{3}{c|}{\centering \includegraphics[width=12mm]{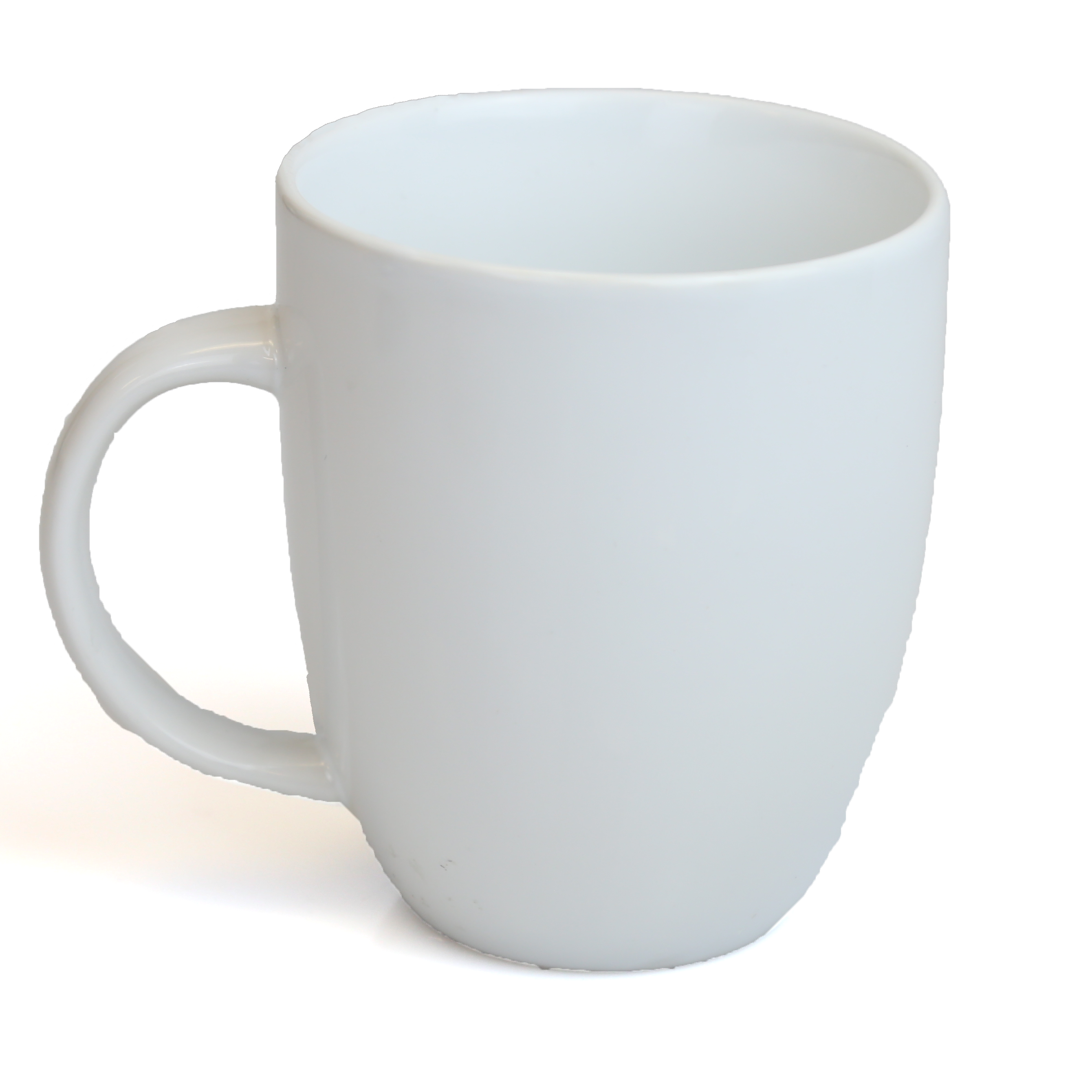}}& \multicolumn{3}{c|}{\centering \includegraphics[width=12mm]{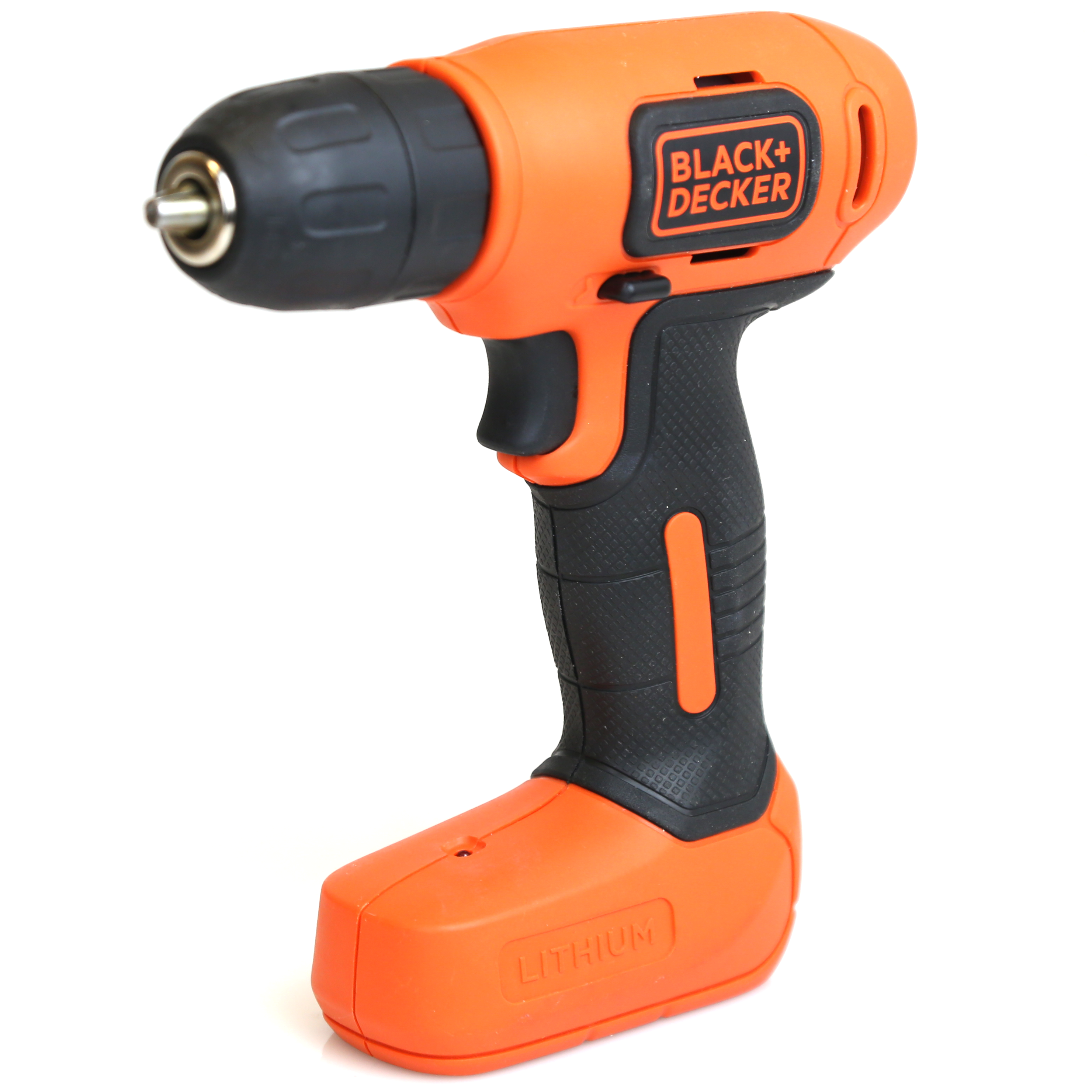}}& \multicolumn{3}{c|}{\centering \includegraphics[width=12mm]{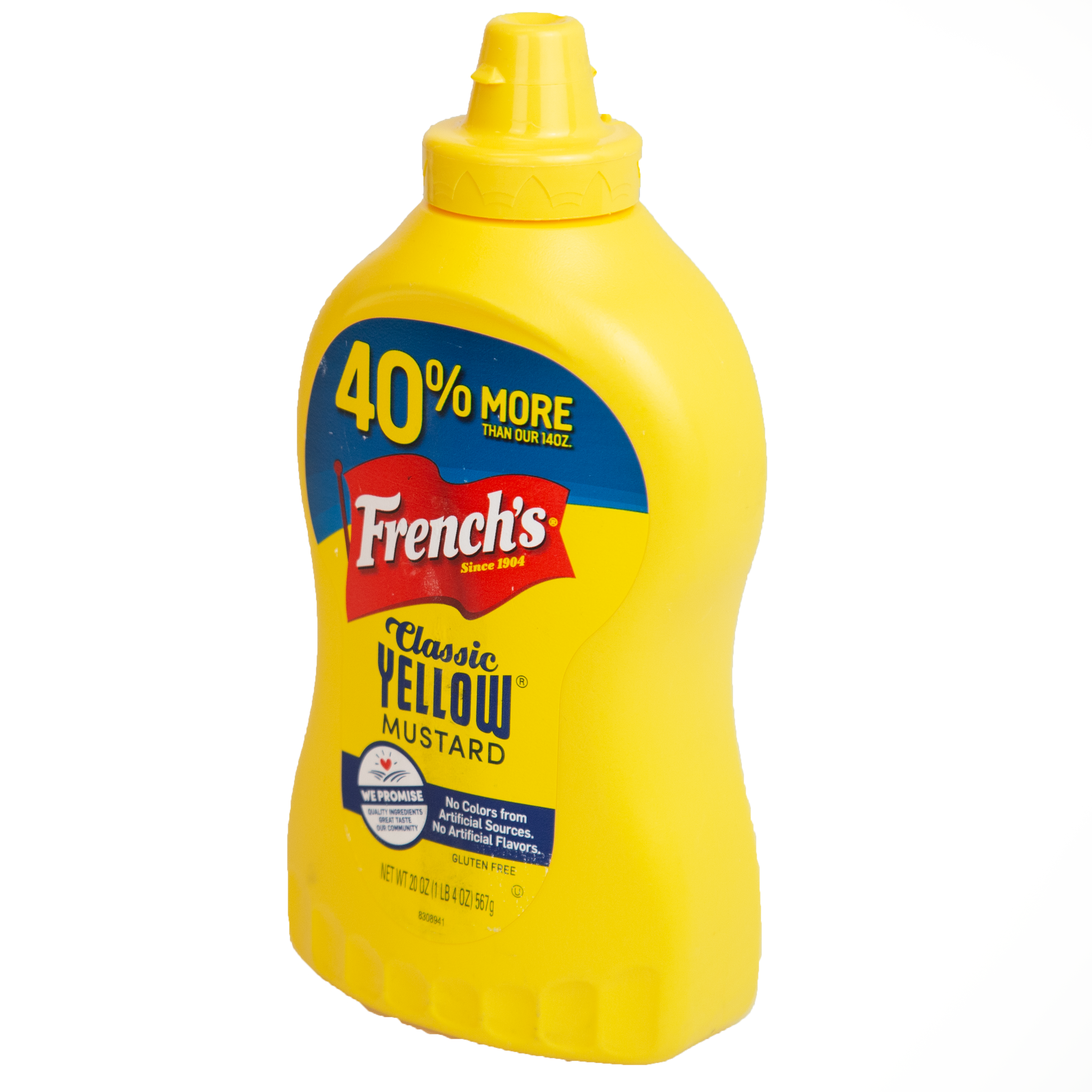}}& \multicolumn{3}{c}{\centering \includegraphics[width=12mm]{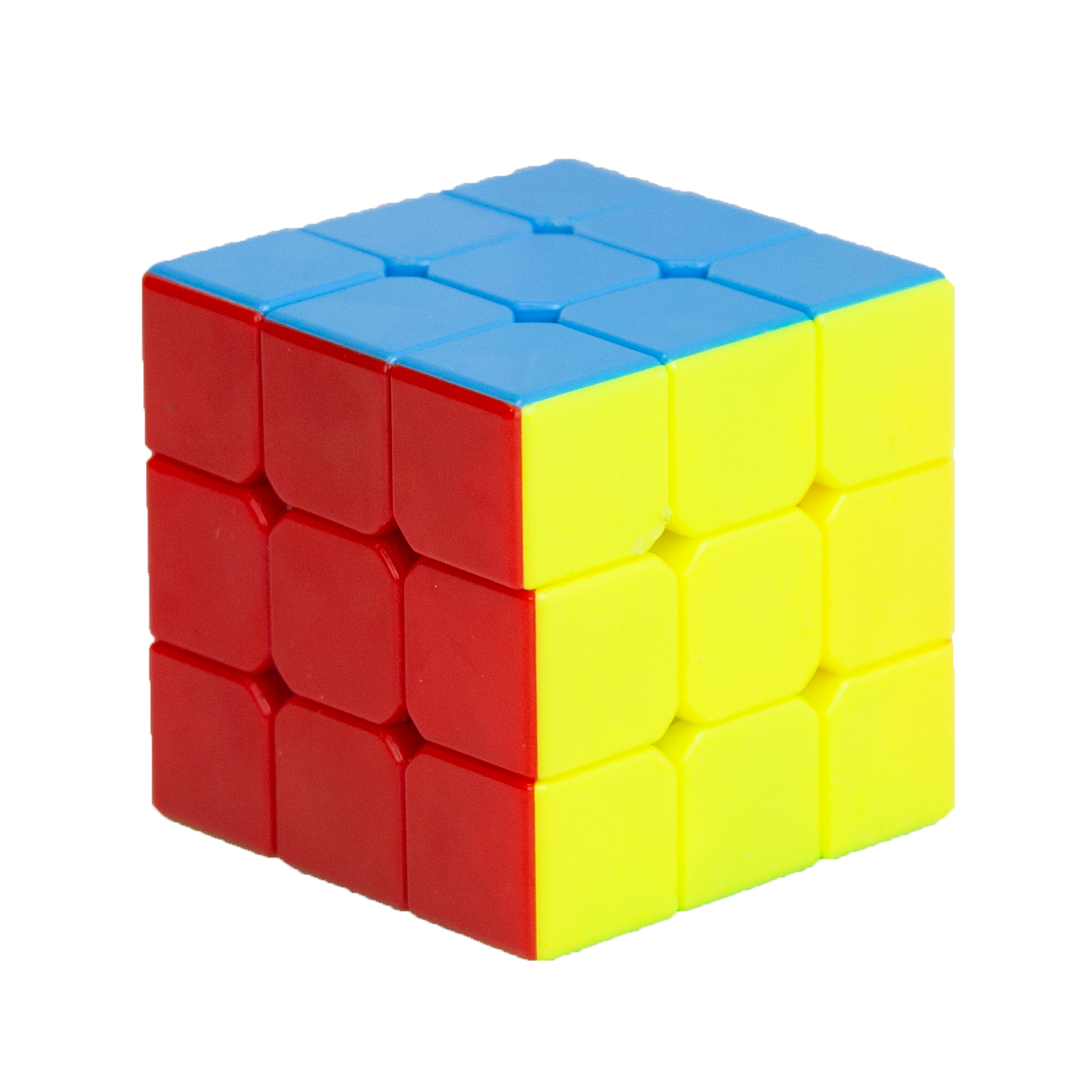}} \\

& \centering Dimensions (mm) & \multicolumn{3}{c|}{218 $\times$ 3 $\times$ 3} & \multicolumn{3}{c|}{156 $\times$ 96 $\times$ 64}& \multicolumn{3}{c|}{135 $\times$ 109 $\times$ 94}& \multicolumn{3}{c|}{175 $\times$ 51 $\times$ 188}& \multicolumn{3}{c|}{92 $\times$ 56 $\times$ 192}& \multicolumn{3}{c}{56 $\times$ 56 $\times$ 56}\\
&\centering Weight (g) & \multicolumn{3}{c|}{197.4} & \multicolumn{3}{c|}{172.6}& \multicolumn{3}{c|}{400.0}& \multicolumn{3}{c|}{613.2 (OOD)}& \multicolumn{3}{c|}{43.4 (OOD)}&\multicolumn{3}{c}{76.8 (OOD)}\\
 \hline
\multirow{5}{10mm}{\newline \newline \newline \newline Prediction Error (mm)}  &\centering
Axis & X & Y & Z & X & Y & Z & X & Y & Z & X & Y & Z & X & Y & Z  & X & Y & Z \\ 
\cline{2-20}
&\centering One Grasp Only& 14.4&17.5&9.4&31.0&9.0&16.2&13.7&\textbf{6.1}&21.1&18.4 &25.3 & 26.4 &25.7&15.3&\textbf{8.1}&17.2&8.4&8.0\\
&\centering Analytical Solution&4.6&\textbf{3.0}&13.7&27.9&4.8&9.3&6.8&6.5&17.3&13.5&\textbf{9.2}&17.9&\textbf{7.3}&8.2&11.4&\textbf{1.8}&4.9&7.5\\
&\centering Random Rotate&15.4&23.2&18.0&26.2&3.8&20.3&11.9&20.9&14.4&\textbf{11.0}&20.1&25.3&25.9&8.5&12.1&15.9&\textbf{4.3}&4.2\\
 &\centering \textbf{U-GRAPH (Ours)}&\textbf{4.5}&7.7&\textbf{8.7}&\textbf{21.6}&\textbf{3.6}&\textbf{8.5}&\textbf{6.7}&6.5&\textbf{10.9}&11.8&15.8&\textbf{15.6}&12.1&\textbf{7.4}&10.1&13.5&9.3&\textbf{3.2}\\

\end{tabular}
\caption{The table shows the mean error of each axis of all 12 real-world objects. We performed 5 different grasp configurations on each object and tried to maximize the variations of $\mathrm{d}x$, $\mathrm{d}y$, and $\mathrm{d}z$ for each grasp. We also show the results of the baseline methods and bold the best-performing estimation for each axis of each object. The $X$, $Y$, and $Z$ axes are defined by the world frame. The OOD label in the last three objects' weight means that their weight is out of our collected data distribution.}
\label{table2}
\vspace{-5mm}
\end{table*}
\subsection{Additional Study on the Effect of Weights of the Object}
\label{add}
The performance of our algorithm is observed to decline when the object weight falls outside the range of our initial data collection, as highlighted in Tab. \ref{table2}. We have set up a focused experiment using the Mustard Bottle (Object 11 in Tab. \ref{table2}) as our primary test subject to further investigate. For this experiment, we supplemented the Mustard Bottle with three different sets of weights, bringing the total weights to 244.6 grams, 446.2 grams, and 648.1 grams, respectively. We tape sets of standard laboratory weights on the side of the mustard bottle around the measured CoM location.
%\yuchen{how}
The first two weights fall within the weight range of our collected dataset, while the last weight surpasses the upper limit of our previous data collection. We maintain a consistent grasping position across all weight variations to isolate the effect of weight on our CoM estimation accuracy. The specific grasping locations and the errors in the CoM predictions made using our method for each weight configuration are documented in Tab. \ref{table3}.

\begin{table}[htbp]
    \normalsize
    \centering
    \begin{tabular}{cm{20mm}|ccc}
     \multirow{6}{*}{\centering \includegraphics[width=25mm]{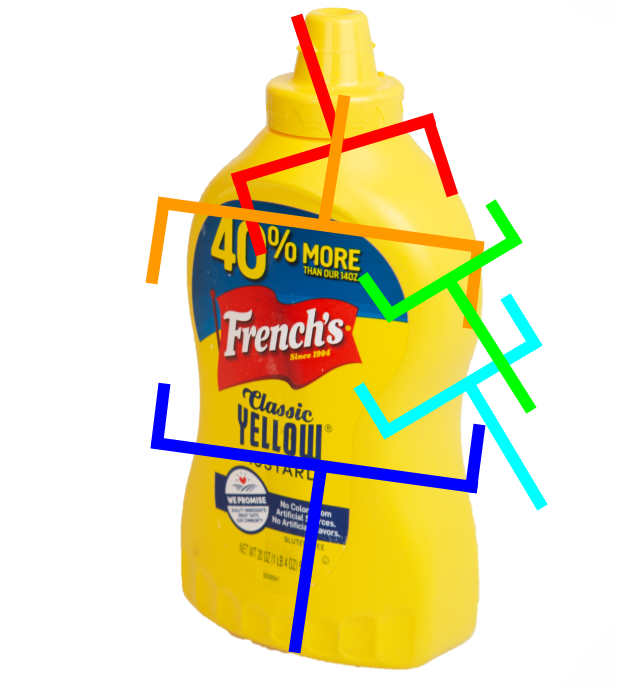}} &  \multirow{2}{=}{\centering Weights (g)} & \multicolumn{3}{c}{Mean Error (mm)}\\
         & & X &Y&Z\\
         \cline{2-5}
         &\centering 43.4 (OOD) & 12.1 & 7.4 & 10.1\\
         &\centering 244.6 & 5.0 & 6.1& 6.4 \\
         &\centering 446.2 & 6.1 &4.4 &5.6 \\
         &\centering 648.1 (OOD) & 15.6 &12.7 & 13.2\\
    \end{tabular}
    \caption{The left image shows the grasping locations on the mustard bottle. The right table shows the mean prediction error on each axis for different weights. The $X$, $Y$, and $Z$ axes are defined by the world coordinate. The OOD label means the weight is out of our training data distribution.}
    \label{table3}
    \vspace{-5mm}
\end{table}
\subsection{Limitation and Discussion}
\label{discuss}

Our approach aims to minimize the total error in CoM estimation, which occasionally compensates for the different axes. As evidenced in Tab. \ref{table2}, while our method might yield slightly poorer results on one axis, it significantly enhances performance on others. However, our method demonstrates superior accuracy in 10 of the 12 test objects compared to other techniques. This demonstrates the effectiveness of taking a second measurement, as highlighted by our comparison with the baseline method, \textbf{One Grasp}, and the importance of informed active perception noted against the \textbf{Random Rotate} method, which lacks the informed approach of our second measurements. We also show that we can always improve the estimation along the z-axis. This aligns with the intuition that a new orientation will introduce new information about the offset of the $z$-axis, with a second action. 

To further expand this work, predictions could benefit from multiple actions and continuous updates to the estimated center of mass. In this paper, our objective is to demonstrate that this predictive framework can enhance CoM estimation, rather than to claim a complete solution to the problem. Future research should investigate the optimal number of rotations and explore whether a series of actions and predictions can converge to the true CoM.
%\yuchen{Update the z-axis discussion}

Sec. \ref{add} illustrates how the objects’ weight range, spanning 127.36 to 585.36 grams, affects our algorithm’s performance. However, in practice, heavier objects are often inherently unstable for grasping. Extremely light objects fail to generate sufficient F/T signals to overcome noise, making it difficult to expand the dataset in these regions. Despite these challenges, our method demonstrates robust performance across diverse real-world objects that differ significantly in contact geometry, surface friction, and density from the training set, confirming its strong generalizability.

Moreover, our system encounters difficulties with large slips, especially with heavier objects or when the CoM is significantly offset from the grasp points. This is a common challenge in achieving stable grasps and accurate CoM estimations. To address these difficulties, we plan to integrate fingertip GelSight sensors \cite{gelsight} into our system in the future. These sensors will enable precise measurement of slips during manipulation, allowing us to gather critical data to refine our algorithm further. By enhancing our ability to detect and adjust for slips, we aim to improve both the stability of grasps and the robustness of CoM estimations.

\section{Conclusions}

This paper presents U-GRAPH, a novel approach to the center of mass estimation with active perception. We design a pipeline that contains two main components: a Bayesian Neural Network that can provide prediction and its associated uncertainty, and an ActiveNet that produces an informed rotation based on the prior estimation. This approach reduces the need for repetitive grasping by replacing it with an efficient and effective rotation. Our experiments validate the effectiveness of U-GRAPH, which consistently outperforms traditional methods and adapts well to real-world scenarios.

\addtolength{\textheight}{0cm}   

\section*{ACKNOWLEDGMENT}

% The authors would like to thank the entire RoboTouch Lab, especially Amin for his help with photos and figures, Harsh for revising the paper, and Xiping for helping make the video.

The authors would like to thank Mohammad Amin Mirzaee for his help with photos and figures, Hung-Jui (Joe) Huang for thoughtful discussions, and Harsh Gupta for revising the paper. We also thank Xiping Sun and Yilong Niu for helping make the video.
% The authors would like to thank Ruohan, Arpit, Jingyi, Dakarai, and Yuchen for their help towards this paper. Special Thanks to Amin for helping with picture making and hardware design.

\bibliographystyle{IEEEtran} 
\bibliography{bibliography.bib}

\end{document}